\documentclass[letterpaper,10pt,conference]{ieeeconf}
\IEEEoverridecommandlockouts

\usepackage{graphicx}
\usepackage{epsfig}
\usepackage{amsmath,amssymb}
\usepackage{booktabs}
\usepackage{multirow}
\usepackage{tabularx}
\let\labelindent\relax
\usepackage{enumitem}
\usepackage{cite}
\usepackage{url}
\usepackage{hyperref}
\hypersetup{hidelinks}
\usepackage{xspace}
\usepackage{algorithm}
\usepackage{algpseudocode}
\usepackage{cleveref}
\usepackage{subcaption}
\usepackage{graphicx}      
\usepackage{tikz}          
\usepackage{pgfplots}

\usepackage{booktabs}
\usepackage{tabularx}
\usepackage{array}
\usepackage{titlesec}
\titlespacing*{\section}      {0pt}{*1}{*0.8}
\titlespacing*{\subsection}   {0pt}{*0.8}{*1.0}
\titlespacing*{\subsubsection}{0pt}{*0.8}{*1.0}

\pgfplotsset{compat=1.18}  

\usepackage{booktabs}      
\usepackage{multirow}      

\usepackage{caption}
\usepackage{subcaption}    

\usepackage{xcolor}

\newcommand{\method}{ManiSkillFormer\xspace}

\newcommand{\SkillGraph}{AgenSkillGraph\xspace}

\setlist[itemize]{leftmargin=*,noitemsep,topsep=2pt}
\setlist[enumerate]{leftmargin=*,noitemsep,topsep=2pt}

\usepackage{listings}
\usepackage{xcolor}
\usepackage{indentfirst}      
\definecolor{oursblue}{RGB}{31,90,160}
\definecolor{a1red}{RGB}{200,80,60}
\definecolor{a2gold}{RGB}{225,160,40}

\lstdefinelanguage{json}{
    basicstyle=\ttfamily\small,
    numbers=none,
    breaklines=true,
    frame=single,
    showstringspaces=false,
    keywordstyle=\color{blue},
    stringstyle=\color{red},
    commentstyle=\color{gray},
}

\usepackage{comment}

\title{\LARGE \bf
\method: Demonstration-Free Compositional Manipulation via Geometric Contracts and Agentic Skill Graph
}

\author{\authorblockN{Peiqi Yu,
Mosam Dabhi,
Shangtao Li, Bowei Li, Laszlo Jeni, and 
Changliu Liu}
\authorblockA{Carnegie Mellon University, Pittsburgh, PA 15213, USA\\ Email: \{peiqiy, mdabhi, shangtal, boweili, laszloaj, cliu6\}@andrew.cmu.edu}
}

\begin{document}



\setcounter{figure}{0}
\twocolumn[{%
\renewcommand\twocolumn[1][]{#1}%
\maketitle


\begin{center}
    \captionsetup{type=figure}
    \includegraphics[width=0.9\linewidth]{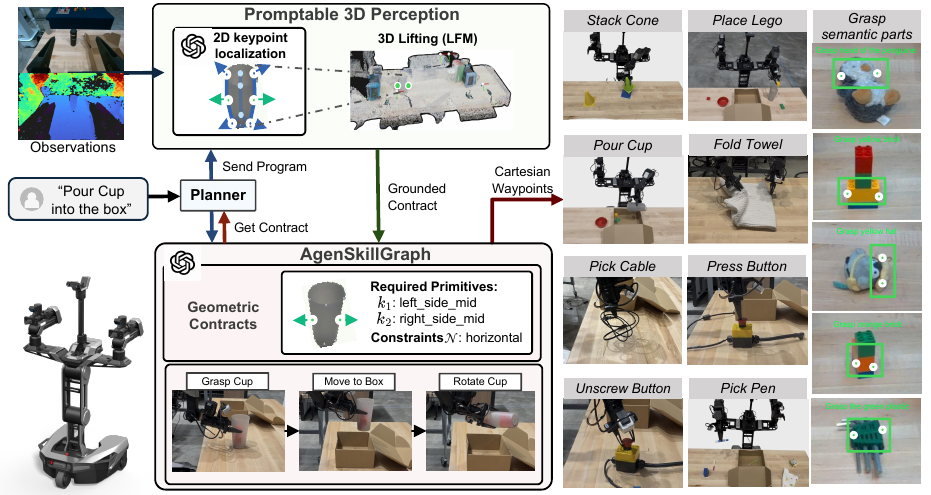}
    \caption{\textbf{ManiSkillFormer online stage overview.} Given a language instruction, the Planner maps the task to structured skill-object pairs and retrieves the corresponding geometric contracts from \SkillGraph. A promptable 3D perception module then grounds the requested semantic 3D primitives from calibrated multi-view observations. The grounded execution packet parameterizes agent-generated motion templates in the SkillGraph to generate Cartesian waypoints. Right side: representative demonstration-free manipulation across different task types.}
    \label{fig:overview}
\end{center}

}]

\thispagestyle{empty}
\pagestyle{empty}

\renewcommand{\footnoterule}{\kern-3pt\hrule width 0.4\columnwidth\kern2.6pt}

\begin{abstract}

Adapting robotic manipulation to new objects and tasks often requires additional demonstrations or manual engineering. Reusable manipulation skills can reduce this effort, but adapting these skills to new scenes remains challenging. We present \method, a framework for demonstration-free and compositional manipulation that connects perception and action through explicit geometric contracts. Building on reusable skill schemas, LLM agents generate contracts specifying the geometry primitives required by each skill, together with corresponding motion templates for semantic objects and task contexts. These contracts guide a perception module to ground task-relevant 3D geometry from observations, which is then used to instantiate reusable motion templates in a skill library.
 We evaluate \method on a dual-arm robot across demonstration-free pick-and-place with 30 instances from 8 object categories, functional manipulation including unscrewing, pouring, pressing, and folding, and three long-horizon tasks. \method achieves an average success rate of 88.97\% for pick-and-place, 75.00\% for functional manipulation, and completion rates of 50--80\% across the long-horizon tasks, outperforming the evaluated baselines and two ablated pipelines. These results demonstrate the potential of explicit geometric contracts to support skill reuse and composition across objects and tasks without per-object policy fine-tuning or additional robot demonstrations. The project website is at \href{https://patricia1019.github.io/ManiSkillFormer/}{https://patricia1019.github.io/ManiSkillFormer/}.

\end{abstract}
\section{INTRODUCTION}
\label{sec:intro}

Robotic manipulation in unstructured environments requires behaviors that generalize across changing objects and compose into long-horizon tasks.
Recent vision-language-action (VLA) models address this challenge by learning end-to-end policies that map observations and instructions directly to actions~\cite{openvla,pi0}.
Although these policies achieve broad task coverage, they require substantial embodied data and computational resources.
More importantly, because perception, task reasoning, and control are entangled within a single learned mapping, it is often difficult to interprete its action selection, identify the source of a failure, or anticipate how the policy will behave in unfamiliar situations.

Hierarchical and modular approaches address this limitation by separating high-level task reasoning from reusable robot skills or programs~\cite{saycan,liang2023code,mishani2025mosaic,li2025nesypackneurosymbolicframeworkbimanual}. This decomposition makes the system more inspectable, but also brings a problem: how do these separated modules communicate with each other in an interpretable, efficient, yet generalizable way?


Let's look at how humans do this. Consider how a human expert might teach a novice to program a robot.
Rather than specifying every Cartesian waypoint, the expert provides mostly semantic guidance: where an object should be grasped and which meaningful intermediate waypoints the robot should pass through.
For example, when instructed to ``grasp the cup to pour,'' the novice may identify two opposing points on the cup's mid-body and select a horizontal approach direction compatible with the subsequent rotation.
The corresponding motion can also be described semantically: move above the cup, approach along the selected direction, and close the gripper, without bringing in exact coordinates.
Once these references are located in the current scene, the semantic description can be instantiated as numerical robot commands.
This process suggests a natural decomposition: 
\textit{the geometric requirements and motion structure of a skill can be represented in semantic space, while only their scene-specific realization and final execution need to be instantiated numerically}.

Inspired by this decomposition, we propose \method, a framework for demonstration-free and compositional robotic manipulation that connects a promptable 3D perception module with a reusable skill library, \SkillGraph, through \emph{task-conditioned semantic geometric contracts}. Each  skill in \SkillGraph pairs a contract specifying what keypoints perception must provide with a semantic skill template specifying how the grounded geometry should become object-relative waypoints. Before deployment, schema-guided agents instantiate  contracts and skill templates. At runtime, a planner retrieves and composes the required skills, the promptable 3D perception module then grounds their requested primitives in the observed scene according to the contract, and the generated skill templates instantiate them as numerical robot waypoints. This design preserves reusable skill requirements while adapting their numerical realization to each scene and object.

Our main contributions are:
\begin{enumerate}
\item \emph{Semantic geometric contracts.}
We introduce an interface letting reusable skills explicitly request the geometry they need from perception, satisfying the needs from both the task side and the skill execution side.

\item \emph{A contract-grounded \SkillGraph.}
We build \SkillGraph with contract-grounded atomic skills and dependency-aware meta skills, semi-autonomously instantiated before deployment by schema-guided agents.

\item \emph{Real-robot validation.}
On the Galaxea R1-Lite, we evaluate generalization to different object instances, functional manipulation, and long-horizon composition.
\end{enumerate}



\section{RELATED WORK}
\label{sec:related}

\subsection{Reusable Skills and Structured Manipulation}

Skill-based manipulation addresses two complementary concerns: organizing reusable behaviors and coordinating their execution.
MOSAIC~\cite{mishani2025mosaic} and NeSyPack~\cite{li2025nesypackneurosymbolicframeworkbimanual} support modular execution through skill libraries and hierarchical abstractions.
Beyond modularity, STAP accounts for geometric dependencies between successive skills using learned feasibility estimates~\cite{stap}.
Building on NeSyPack's SkillGraph, we compose basic actions (atomic skills) into reusable sequences (meta skills), providing a compact, semantically meaningful vocabulary for LLM planning.

\subsection{Task-Relevant Geometric Perception}
Recent VLMs support language-conditioned 2D point prediction~\cite{molmo,robopoint}, making them suitable for localizing the interaction points described by geometric contracts.
Using these predictions for manipulation requires recovering their 3D geometry, but direct back-projection relies on depth measurements that may be incomplete or noisy.
Lifting foundation models complement 2D localization by inferring 3D structure from landmarks across object categories~\cite{2dlfm}.
Thus, we combine VLM 2D landmark prediction with a SoTA 3D lifting model to reduce reliance on incomplete or noisy depth measurements.

\subsection{Perception--Action Interfaces}

Previous interfaces connect perception to action through constraints or learned policies.
Explicit approaches guide motion generation using value maps, as in VoxPoser~\cite{voxposer}, or geometric constraints, as in ReKep, CoPa, and GeoManip~\cite{rekep,copa,geomanip}.
MOKA and OmniManip express interactions through visual marks and object-centric primitives~\cite{moka,pan2025omnimanip}.
Learned approaches instead condition execution on keypoints or spatial primitives~\cite{kite,kalm,hacmanpp}.
Our contracts provide an explicit interface specifying the task-dependent interaction geometry required by reusable motion templates.

\subsection{LLM-Based and Agentic Manipulation}

LLM-based frameworks organize robot behavior at different levels of abstraction.
At the skill level, SayCan selects actions using learned affordances~\cite{saycan}, while RoboMatrix combines task decomposition with execution checking over learned meta-skills~\cite{mao2024robomatrix}.
At the program level, Code as Policies generates executable logic over perception and control APIs~\cite{liang2023code}.
Our agents focus on skill adaptation, tailoring human-defined skills to different objects and task contexts by jointly specifying what to perceive and how to move.
\begin{figure*}[t]
    \centering
    \includegraphics[width=0.9\textwidth]
    {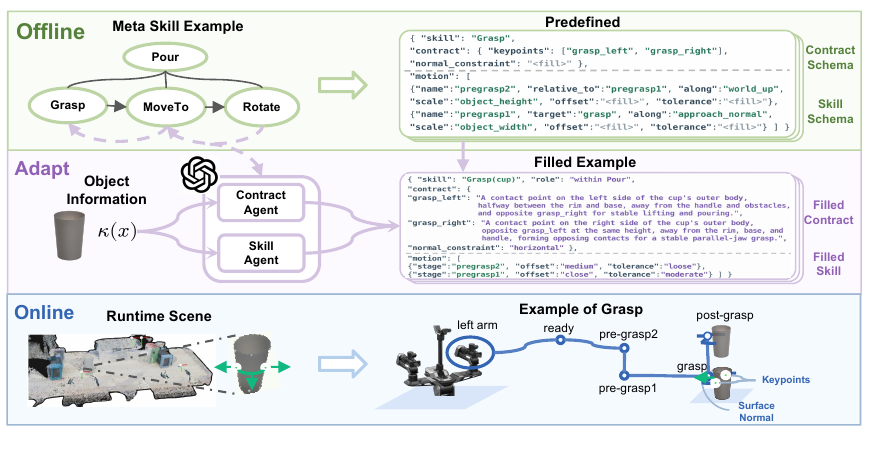}
    \vspace{-0.5em}
    \caption{Overview of the Agentic SkillGraph pipeline. Offline, human experts define reusable skill structurs and contract schemas. During adaptation, two LLM agents tailor these schemas to each semantic skill--object pair and its task context, generating geometric contracts and semantic motion templates. Online, the perception module grounds the requested geometry in the observed scene, which parameterizes the motion templates for robot execution. Green denotes human-defined fields, and purple denotes agent-generated fields.}
    \label{fig:agenskillgraph}
    \vspace{-1em}
\end{figure*}

\section{PROBLEM FORMULATION}
\label{sec:problem}
We consider tabletop lightweight manipulation: a dual-arm manipulator with
parallel-jaw grippers acting on rigid or mildly deformable objects that rest
within reach on a planar work surface. 
The action vocabulary is constrained: manipulation verbs (skills) are drawn from a fixed
set $\mathcal{V}$ of tabletop manipulation verbs, chosen to be representative of
common tabletop tasks. Object nouns are
open-vocabulary.

To formalize the problem, we introduce three spaces:
\begin{itemize}
  \item \textbf{Semantic space $\mathcal{L}$} contains user instructions and semantic relationship between objects and skills, which is represented and reasoned in the language space. 
  \item \textbf{Observation space $\mathcal{O}$} contains raw robot observation and its scene-specific geometric grounding such as  3D keypoints and surface normals.
  \item \textbf{Numerical space $\mathcal{N}$} contains robot execution parameters, including waypoints in Cartesian space and joint space $\tau_i$.
\end{itemize}

The robot receives an instruction $l\in\mathcal{L}$ and an initial observation $o_1\in\mathcal{O}$. We assume that $l$ specifies an ordered sequence of skill calls, each involving at most one in-hand object and one target object: 
\begin{equation} l= \big((s_1,\mathbf{X}_1),\ldots,(s_H,\mathbf{X}_H)\big) \in\mathcal{L}, \quad \mathbf{X}_i=(x_i^{\mathrm{h}},x_i^{\mathrm{t}})\label{eq:program} 
\end{equation} 
where $s_i\in\mathcal{V}\subseteq\mathcal{L}$ is a symbolic skill call, and $x_i^{\mathrm{h}},x_i^{\mathrm{t}} \in\mathcal{X}\cup\{\varnothing\}$ are symbolic arguments specifying the in-hand and target objects respectively, with $\mathcal{X}\subseteq\mathcal{L}$. 

Executing each call requires grounding its symbolic object arguments in the observation space and its symbolic skill in the robot space. We denote the grounded object arguments by $\hat{\mathbf{X}}_i\in\mathcal{O}$ and the grounded skill by $\tau_i=\hat{s}_i(\theta_i,\hat{\mathbf{X}}_i)$, where $\theta_i\in\mathcal{N}$ denotes its skill parameters, $\hat{s}_i$ denotes its skill policy, and $\tau_i$ denotes the final trajectory segment in the robot joint space.

For example, consider a call that picks up a mug: $
    (s_i,\mathbf{X}_i)
    =
    (\texttt{Pick},(\texttt{mug},\varnothing)) \in \mathcal{L}.
$
The symbolic argument \texttt{mug} is first grounded to the observed mug
$\hat{\mathbf{X}}_i \in \mathcal{O}$. Based on its observed geometry, $\theta_i$ may specify grasp pose, approach direction, and lift distance in the numerical space as grounded execution parameters.  
Based on $\theta_i$ and $\mathbf{X}_i$ , The skill policy $\hat{s}_i$ then generates the trajectory segment
$
    \tau_i\in\mathcal{N},
$
which specifies the sequence of robot motions used to execute the call.

In general, given $(u,o_1)$, the objective is to ground each symbolic call $(a_i,\mathbf{X}_i)$ and produce an executable trajectory sequence $\tau=(\tau_1,\ldots,\tau_H)$ to complete the instruction $l$.


\begin{algorithm}[t]
\caption{Online execution of \method}
\label{alg:execution}
\begin{algorithmic}[1]
\Require instruction $l$,  observation $o_0$, \SkillGraph $\mathcal{G}$, promptable 3D perception module, robot controller $\mathcal{C}_r$
\Ensure execution success or failure
\State $\Phi \gets \mathrm{InitState}(o)$
\State $p \gets \mathrm{Plan}(l;\mathcal{G})$ \Comment{via $\mathcal{D}$}
\State $c_p \gets \mathrm{RetrieveContract}(p;\mathcal{G})$ 
\State $\boldsymbol\theta \gets Perception(o,c_p)$ \Comment{ground requested primitives}
\State $\hat p \gets ((a_i,x_i,\theta_i))_{i=1}^{N}$
\For{$i=1$ to $N$}
    \State $(a_i,x_i,\theta_i) \gets \hat p[i]$
    \If{$\neg\,\mathrm{Pre}_{a_i}(x_i,\Phi)$} \Return \textbf{False} \EndIf
    \State $\mu_i \gets \mathrm{RetrieveTemplate}(a_i,\kappa(x_i);\mathcal{G})$ 
    \State $\tau_i \gets \mu_i(\theta_i)$
    \State $\mathcal{C}_r(\tau_i)$
    \State $\Phi \gets \mathrm{UpdateState}(\Phi,a_i,x_i)$
    \If{$\neg\,\mathrm{Post}_{a_i}(x_i,\Phi)$} \Return \textbf{False} \EndIf
\EndFor
\State \Return \textbf{True}
\end{algorithmic}
\end{algorithm}

\section{METHOD}
\label{sec:method}
\subsection{System Overview}
\label{sec:overview}
As shown in \Cref{fig:overview}, method has four components: a rule-based \textit{planner} that parses instructions into skill--object calls; a promptable 3D perception module that grounds a skill's geometric requirements on the observed object; a skill library \textit{\SkillGraph} $\mathcal{G}$ storing reusable skill definitions; and an embodiment-dependent \textit{controller}.
The promptable 3D perception module and \SkillGraph are connected through task-conditioned geometric contracts.
The framework operates in three stages (\Cref{fig:agenskillgraph}): offline schema defining; semi-online skill adaptation; and online execution. 

\textbf{Offline Stage---Schema Defining.}
In the offline stage, we need to prepare an \SkillGraph for the specific task.  

To prepare a \SkillGraph $\mathcal{G}$, we (as human experts) first define semantic atomic skills $\mathcal{A}$, meta skills $\mathcal{M}$, and decomposition rules $\mathcal{D}$, where \(\mathcal{D}\) specify how atomic skills are organized to realize each meta-skill.
Then we should define a contract schema $\Sigma^c_a$ and a skill schema $\Sigma_a$ for each atomic skill $a\in\mathcal{A}$. The contract schema $\Sigma^c_a$ specifies what primitive labels a skill might require for perception module to ground, and leave blank for agent in semi-online stage to fill in the semantic descriptions for these primitives and approach normal constraints (see Sec.~\ref{sec:contract} for details). 
The skill schema $\Sigma_a$ fixes the ordered motion stage
of $a$ in semantic space (see Sec.~\ref{sec:skillgraph} for details). 

These design are in semantic space $\mathcal{L}$, maintaining human knowledge in a easily interpretable and configurable way but are irrelevant to specific scene, reducing manual scene-by-scene skill specification while constraining subsequent agent implementation to remain interpretable and consistent. 

\textbf{Semi-online---Plan and Skill Adaptation.}
Given instruction $l$, planner first plans the skill-object sequence:
\begin{equation}
p = \mathrm{Plan}(l;\mathcal{G}) = ((s_1,x_1),\ldots,(s_k,x_k))\in\mathcal{L}
\label{eq:plan}
\end{equation}

Each skill $s_i$ is either atomic or meta. For an atomic skill, two LLM agents generate the contract and skill template for the skill-object pair. A meta skill 
is first expanded into its constituent atomic skills $a_i$, after which the agents generate a contract and skill template for each atomic skill, conditioned on its role within the meta skill (see Sec.~\ref{sec:contract} for details):
\begin{equation}
\begin{aligned}
c_{a_i,\kappa(x_i)} &= \mathrm{FillContract}(\Sigma^c_{a_i},\Sigma_{a_i},\kappa(x_i))\in\mathcal{L},\\
\mu_{a_i,\kappa(x_i)} &= \mathrm{FillSkill}(\Sigma_{a_i},c_{a_i,\kappa(x_i)},\kappa(x_i))\in\mathcal{L}
\end{aligned}
\label{eq:fill}
\end{equation}
$x_i$ is the semantic object name, 
like ``grey toy"; $\kappa(x_i)$ is the object category that can be specified by human, including \texttt{rigidity}, \texttt{size}, and \texttt{shape}. If not specified, the agent would generate purely according to semantic object name $x_i$;
$c_{a,\kappa(x)}$ is the filled contract which states what perception must find; $\mu_{a,\kappa(x)}$ is the filled skill template states how the grounded result becomes Cartesian waypoints; both remain free of scene coordinates. 
This stage yields an atomic skill execution program $p^{c}=((a_i,x_i,c_{a_i,\kappa(x_i)},\mu_{a_i,\kappa(x_i)}))_{i=1}^{N}$, attach each step with a filled contract and skill template. In this stage, the filled results are also reasoned in semantic space $\mathcal{L}$, utilizing LLM's world knowledge.

\textbf{Online---Ground and Execute.}
Given observation $o$ and the filled program $p^{c}$ from the semi-online stage, \method performs
\begin{equation}
\hat{p} = \mathrm{Ground}(o,p^{c})\in\mathcal{O}, \quad \tau = \mathrm{Execute}(\hat{p};\mathcal{G})\in\mathcal{N}
\label{eq:pipeline}
\end{equation}
$\mathrm{Ground}$ is done by the promptable 3D perception module, replacing each contract in $p^{c}$ with its grounded realization $\theta_i$, producing $\hat{p}=((a_i,x_i,\theta_i,\mu_{a_i,\kappa(x_i)}))_{i=1}^{N}$.
$\mathrm{Execute}$ instantiates each trajectory segment $\tau_i=\mu_{a_i,\kappa(x)_i}(\theta_i)$, which is then dispatched to controller $\mathcal{C}_r$.

The below sections would go into module details.

\subsection{Planner}
\label{sec:planner}
Given instruction $l$, the planner uses a rule-based parser to extract an ordered sequence of skill--object calls by matching each verb phrase to the skill vocabulary $\mathcal{A}\cup\mathcal{M}$ and its adj-noun pair phrases to object arguments, ordered by clause sequence in $l$.
For example, ``pick red lego and place into black cup, then pour black cup into brown box'' yields $
\{l:\texttt{Pick}(\texttt{red lego},\varnothing)$, $\texttt{Place}(\texttt{red lego},\texttt{black cup})$, $\texttt{Pour}(\texttt{black cup},\texttt{brown box})\}
$. 
The final result is a skill sequence $p=((a_1,x_1),\ldots,(a_N,x_N))$.

\subsection{Task-Conditioned Geometric Contracts}
\label{sec:contract}
A geometric contract $c_{a,\kappa(x)}$ specifies what information perception must provide for skill execution.

\textbf{Semantic Keypoints and Surface Normal Constraints.}
The contract should contain keypoint request $k_{a,\kappa(x)}={(\ell_a^i,d^i_{a,\kappa(x)})}_{i=1}^N$: $N$ keypoints, each with a semantic label $\ell_a$ and a free-form description $d_{a,\kappa(x)}$ of where it lies on the object, e.g., \textit{``center point on one well-exposed side of the bottle's main body''}.
It also contains a surface normal constraint
$n_{a,\kappa(x)}$, which specifies the admissible surface
orientation at the interaction region in the robot base frame.
The constraint can specify an angular range relative to
a world-frame axis or plane, such as $30^\circ$--$60^\circ$
relative to the vertical $z$-axis.
The labels \texttt{upright} and \texttt{horizontal}
provide shorthand for alignment with the vertical
$z$-axis and the plane orthogonal to $z$, respectively,
within a specified angular tolerance.
If the constraint is \texttt{none}, no surface
orientation restriction is imposed.  At runtime, the 3D perception module jointly grounds each requested keypoint and its local surface normal, selecting candidate locations whose estimated normals lie within the admissible set \(\mathcal{N}_{a,\kappa(x)}\). 
If the constraint is \texttt{none}, no orientation class is imposed. This surface normal would decide end-effector approach direction. For a parallel-jaw grasp, the grounded contact points and the surface normal would help determine the 6-DoF contact pose used to instantiate the skill template.

\textbf{Example.}
As shown in \Cref{fig:agenskillgraph}, the contract schema $\Sigma^c_a$ should explicitly specifies the semantic keypoints name $\ell_a$ needed for skill execution in advance (e.g., \texttt{grasp\_left}/\texttt{grasp\_right} for \texttt{Grasp}, shown in \textcolor{green!50!black}{green}).
A contract agent (gpt-5.6-sol~\cite{openai_gpt56sol}) then fills $d_{a,\kappa(x)}$ and $n_{a,\kappa(x)}$ (\textcolor{purple!50!black}{purple}).
We assume a parallel-jaw gripper, so \texttt{Grasp} requests two opposing keypoints; other grippers would require a different keypoint set.

Consider $\texttt{Grasp(cup)}$ within $\texttt{Pour}(\text{cup},\text{box})=\texttt{Grasp}\to\texttt{MoveTo}\to\texttt{Rotate}$.
Taken alone, $\texttt{Grasp(cup)}$ admits either class: a top-down grasp on the rim, whose contact normals are \texttt{upright}, or a side grasp on the body, whose normals are \texttt{horizontal}.
The downstream \texttt{Rotate} eliminates the first, since a rim grasp would cause self collision in the subsequent \texttt{Rotate} skill.
The agent therefore sets $n_{a,\kappa(x)}=\texttt{horizontal}$ and requests two opposing points on the body wall.

\subsection{Contract-Guided Promptable 3D Perception Module}
\label{sec:perceim}
This module maps observations and a contract request to grounded object-centric 3D
primitives in Cartesian space.

\textbf{Input.}
This module takes an observation and a contract request. The observation is a set of
calibrated camera views. Our platform provides two RGB-D and two RGB views, and
this module runs on any subset containing depth or at least two views. The contract request contains
the keypoint set $\mathcal{K}_{a,\kappa(x)}$ in \Cref{sec:contract}, and the skill $a$ and object $x$ it belongs to.

\textbf{Task-conditioned 2D Keypoint Prediction.}
We use gpt-5.6-sol to predict
semantic 2D interaction points from RGB images.
For each view, we construct a textual query specifying the
skill, target object, keypoint labels and their descriptions
$\{(\ell_a^i,d_{a,\kappa(x)}^i)\}_i$ required by the contract. Requested keypoints are queried jointly, and the model
returns image-plane coordinates associated with their
semantic labels. These labeled predictions are passed
to a 3D lifting model described below to obtain
the 3D keypoints used to parameterize the motion templates.
No task-specific fine-tuning is required in this stage.

\textbf{3D Lifting and Constraint Checking.}
Each predicted 2D keypoint is lifted to 3D from the fused
observations using the LFM 3D lifting model~\cite{2dlfm}.
For each candidate set, we estimate the local surface
normal $\mathbf{n}$ at the interaction region and express
it in the robot base frame.
We then compute its angle relative to the axis or plane
specified by the contract and check whether this angle
falls within the prescribed range.
Only candidate sets satisfying
$\mathbf{n}\in\mathcal{N}_{a,\kappa(x)}$ are retained
for motion parameterization.
If the constraint is \texttt{none}, this check is skipped.

\textbf{Robot Frame Grounding.}
Using the calibrated camera intrinsics and extrinsics,
we express each reconstructed 3D keypoint in the robot
base frame. For robot-mounted cameras, we compute the
camera pose at capture time using the robot's forward
kinematics and current joint configuration, accounting
for changes in camera pose across observations.

\subsection{\SkillGraph: Agent-Assisted Skill Construction}
\label{sec:skillgraph}

\SkillGraph stores human-defined skill schemas and the agent-generated parameterized skills.

\textbf{Skill Schema.} 
Inspired by how experts teach novices, $\Sigma_a$ encodes human knowledge as an ordered sequence of semantic motion stages. The stages are chosen based on changes in contact mode or relative robot--object pose; for example, \texttt{Grasp} follows $\texttt{pre-grasp2}\to\texttt{pre-grasp1}\to\texttt{grasp}\to\texttt{post-grasp}\to\texttt{retract}$ (see the example in \Cref{fig:agenskillgraph}). In view that how the novice implement these semantic waypoints should adapt to the size of objects it operates on, each stage specifies an \emph{offset distance} (\texttt{close}, \texttt{medium}, or \texttt{far}). The offset is normalized by a \emph{scale reference}, namely a task-relevant object dimension such as its width or height, and is converted into a category-dependent numerical displacement during adaptation (Sec.~\ref{sec:motion_template}). Also, not all waypoints should be strictly reached, so we specify a \emph{reaching tolerance} (\texttt{precise}, \texttt{moderate}, or \texttt{loose}), allowing contact-critical stages to remain precise while intermediate stages serve as flexible guidance.
 
In addition, $\Sigma_a$ defines $\chi_a=(\mathrm{Pre}_a,\mathrm{Post}_a)$, whose conditions are evaluated by internal utility functions, such as checking whether the gripper is open or the object is in hand. As shown in Algorithm~\ref{alg:execution}, $\mathrm{Pre}_a$ must hold before the skill is attempted and $\mathrm{Post}_a$ after its execution; failure of either check terminates execution.

\textbf{Filling a Motion Template.}\label{sec:motion_template}
During adaptation, given a robot reachibility constraint, the skill agent (gpt-5.6-sol) 
assigns each waypoint an \textit{offset-distance} class and a \textit{reaching-tolerance} class.
Each offset-distance class maps to a range expressed as a fraction of the stage's scale reference (e.g., \texttt{close} is $0.15$--$0.35\times$ the reference); each reaching-tolerance class maps to a fixed absolute range, used for execution to decide how far a waypoint may deviate from its nominal pose.  
These classes are generated and resolved against the object the skill operates on, which is what lets a single $\mu_{a,\kappa}$ generalize across instances of different size and shape. \Cref{fig:agenskillgraph} shows an example of the motion templates of $\Sigma_{\texttt{Grasp}}$, what skill schema pre-defined is colored in \textcolor{green!50!black}{green}, and what filled by agent is colored in \textcolor{purple!50!black}{purple}.

\textbf{Atomic and Meta Skills.}
\begin{table}[t]
\centering
\caption{Atomic Skills and Meta-Skill Decompositions.}
\label{tab:skills}
\vspace{-0.4em}

\resizebox{\columnwidth}{!}{%
\begin{tabular}{@{}llll@{}}
\toprule
\multicolumn{2}{c}{Atomic Skills}
&
\multicolumn{2}{c}{Meta Skills} \\
\cmidrule(lr){1-2}
\cmidrule(lr){3-4}
Contact Mode & Atomic Skill & Meta Skill & Skill Decomposition \\
\midrule
No contact
& \texttt{MoveTo}
& \texttt{Pick}
& \texttt{Grasp} $\rightarrow$ \texttt{MoveTo} \\

Establish contact
& \texttt{Grasp}
& \texttt{Place}
& \texttt{MoveTo} $\rightarrow$ \texttt{Release} \\

Maintain contact
& \texttt{Rotate}
& \texttt{Pour}
& \texttt{Grasp} $\rightarrow$ \texttt{MoveTo}
  $\rightarrow$ \texttt{Rotate} \\

Terminate contact
& \texttt{Release}
& \texttt{Unscrew}
& \texttt{Grasp} $\rightarrow$ \texttt{Rotate} \\

Interaction under contact
& \texttt{Press}
& \texttt{Fold}
& \texttt{Press} $\rightarrow$ \texttt{Grasp}
  $\rightarrow$ \texttt{MoveTo} \\
\bottomrule
\end{tabular}%
}

\vspace{-2em}
\end{table}
As shown in \Cref{tab:skills}, we organize the atomic skill set by \emph{contact mode}, making them the minimal units that are reusable yet interpretable in this paper. Meta skills are compositions of existing atomic skills. We present the meta skills used in this paper in \Cref{tab:skills}. These meta skills were chosen because they span different contact-transition sequences, are reusable across tasks, and together cover the functional manipulation and long-horizon tasks in our experiments. Decomposition rules $\mathcal{D}$ are predefined; 
this keeps the atomic sequence for a given meta skill consistent and inspectable.
Extending $\mathcal{G}$ to additional meta skills requires simply writing a new decomposition rule.

\subsection{Controller and Execution}
\label{sec:execution}
Online, the perception module grounds the object parameters
in $\theta_i$. Each stage's offset-distance range is scaled
by the object dimensions, after which an offset and a tracking
tolerance are uniformly sampled from their respective ranges.
This yields Cartesian waypoints $X_i^{(j)}$ with tracking
tolerances $\epsilon_i^{(j)}$.
The controller $\mathcal{C}_r$ assigns arms based on reachability.
For dual-arm skills, the template fixes inter-arm coordination,
so selecting the arm for the first action determines all
subsequent arm assignments. Using the waypoints and tolerances,
the controller solves inverse kinematics subject to
self-collision avoidance, generates the joint-space trajectory
$\tau_i$, and executes it via low-level PD control.
\section{EXPERIMENTS}
\label{sec:experiments}

We evaluate \method by three research questions:

\textbf{Q1 (Object generalization).}
Can skills generalize to different object instances without further demonstrations?

\textbf{Q2 (Functional manipulation).}
Does skill-specific contract return the right geometry, including when the same object is used by different skills?

\textbf{Q3 (Long-horizon composition).}
Can we compose skills for long-horizon tasks with task dependencies reliably?


\subsection{Experimental Setup}
\textbf{Robots and Sensing.}
Evaluations are conducted on \textit{Galaxea R1-Lite}, a dual-arm mobile manipulator with wrist RGB-D and head cameras and a parallel-jaw gripper.

\textbf{Baseline and Ablation Study.} We compare against two external baselines and two ablations.
\textbf{MOKA}~\cite{moka} uses mark-based visual prompting to predict point-based affordances and generate manipulation motions, providing a comparison with a task-conditioned geometric interface.
\textbf{VLA system} $\pi_0$~\cite{pi0} is fine-tuned with 50 demonstrations per object category and evaluated on pick-and-place.
\textbf{A1 (no contract)} passes the task description directly to the promptable 3D perception module, which returns the number of keypoints required by the skill without explicit contract-defined semantic requests.
\textbf{A2 (independent generation)} generates each atomic skill's contract and motion template independently, without conditioning on subsequent skills in the meta skill.
All methods use the same object instances, layouts, task instructions, and success criteria.


\textbf{Task Suites.}
We organize the evaluations into three tabletop suites:
1) The \emph{instance-transfer suite} evaluates pick-and-place across 8 object groups comprising 30 instances in total, with 17 trials per group.
2) The \emph{functional manipulation suite} evaluates unscrewing, pouring, pressing, and folding, with 17 trials per task, to examine geometric grounding required by different skills.
3) The \emph{long-horizon compositional suite} evaluates three multi-step tasks, with 10 trials per task, to examine skill composition and dependencies between successive actions.

\textbf{Metric.}
A trial is successful if the robot achieves the task goal.
We report success rates averaged across repetitions.

\subsection{Results}

\subsubsection{Q1 (Object Generalization)}
Table~\ref{tab:object_pick_place} reports single-object pick-and-place performance across 8 object groups with 30 instances in total (136 trials for each method). According to the results, \method achieves the highest or joint-highest success rate in seven of the eight object groups, supporting the robustness of contract generation and the reuse of generated motion templates in demonstration-free cases without per-instance tuning.

\begin{table}[htbp]
\centering
\vspace{-0.5em}
\caption{Single-object pick-and-place success rates (\%).}
\label{tab:object_pick_place}
\vspace{-0.5em}
\setlength{\tabcolsep}{3pt}
\begin{tabular}{lccccc}
\toprule
Object Group & \method & MOKA & A1 & A2 & $\pi_0$ \\
\midrule
Bottle / Cup   & \textbf{94.12} & 64.71 & 76.47 & 88.24 & 41.18 \\
Pen            & \underline{\textbf{88.24}} & 41.18 & 70.59 & \underline{\textbf{88.24}} & 35.29 \\
Cables / Towels & \textbf{94.12} & 29.41 & \textbf{94.12} & 76.47 & 29.41 \\
Bowl           & \underline{\textbf{100}}   & 64.71 & 88.24 & \underline{\textbf{100}}   & 82.35 \\
Toys           & \underline{\textbf{88.24}} & 58.82 & 76.47 & \underline{\textbf{88.24}} & 64.71 \\
Tools          & 82.35         & 47.06 & \textbf{94.12} & 76.47 & 35.29 \\
Cone           & \textbf{94.12} & 64.71 & 64.71 & 70.59 & 41.18 \\
Lego Structures & \textbf{70.59} & 47.06 & 58.82 & 58.82 & 35.29 \\
\midrule
Average & \textbf{88.97} & 52.21 & 77.94 & 80.88 & 45.59 \\
Standard Deviation      & \textbf{9.13} & 13.13 & 13.25 & 12.87 & 18.27 \\

\bottomrule
\end{tabular}
\vspace{-0.5em}
\end{table}

\begin{figure}
    \centering
    \includegraphics[width=1.01\linewidth]{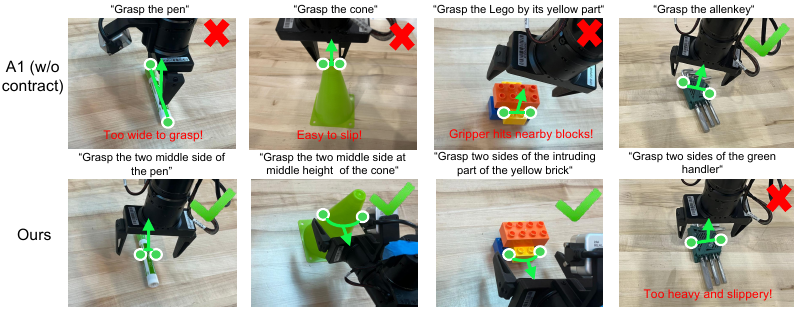}
    \caption{Comparison of A1 (no contract) and \method}
    \label{fig:comparison-A1}
    \vspace{-2em}
\end{figure}

\textbf{Where contracts help most.}
 \method outperforms MOKA and $\pi_0$ on all eight object groups. Compared with A1 (no contract), \method achieves the largest gains on cones (29.41\%), followed by pens (17.65\%).
As illustrated in \Cref{fig:comparison-A1}, without an explicit contract specifying \emph{where} to grasp, perception can return distinctive edge or corner landmarks that are unsuitable for stable execution. The reason why A2 (independent generation) performs very close to the full pipeline is that for this task suite, we only test on \texttt{Pick} and \texttt{Place}, which do not have much geometry dependencies temporally.

\textbf{Semantic Grounding}
Lego structures and toys illustrate how semantics help localize keypoints.
A task instruction $l$ such as ``pick the Lego by the yellow part'' or ``pick the penguine by its head'' names a specific semantic target. 
\method's \texttt{Pick} contract request guides the promptable 3D perceptiom module to extract the exact semantic parts that meet both the task requirement and the skill execution requirement; the A1 ablation, lacking this explicit request, often returns less constraint keypoints that is hard for robot to execute.
$\pi_0$ fails here because with only 50 demos per category the learned policy does not generalize to specific semantic request or unseen configurations. 

\textbf{Where contracts do not help.}
\method underperforms the A1 ablation on \emph{Tools} (82.35 vs.\ 94.12).
For Tools like allenkey, the contract requests a grasp at the handle center, and because the surface of the handle is slippery, and the allenkey has a relatively heavy weight compared to the gripper itself, the robot would drop the allenkey after picking up. But for the A1 ablation, it returns two visually salient keypoints at the edge of the handle, which unexpectedlly forms a more stable grasp.



\subsubsection{Q2 (Functional Manipulation)}

Table~\ref{tab:functional} evaluates functional manipulation beyond pick-and-place, including \emph{unscrew}, \emph{pour}, \emph{press}, and \emph{fold}.
These tasks examine both skill-specific geometric grounding and the coordination of atomic skills within meta skills.

\begin{figure}[h]
    \vspace{-1em}
    \centering
    \includegraphics[width=\linewidth]{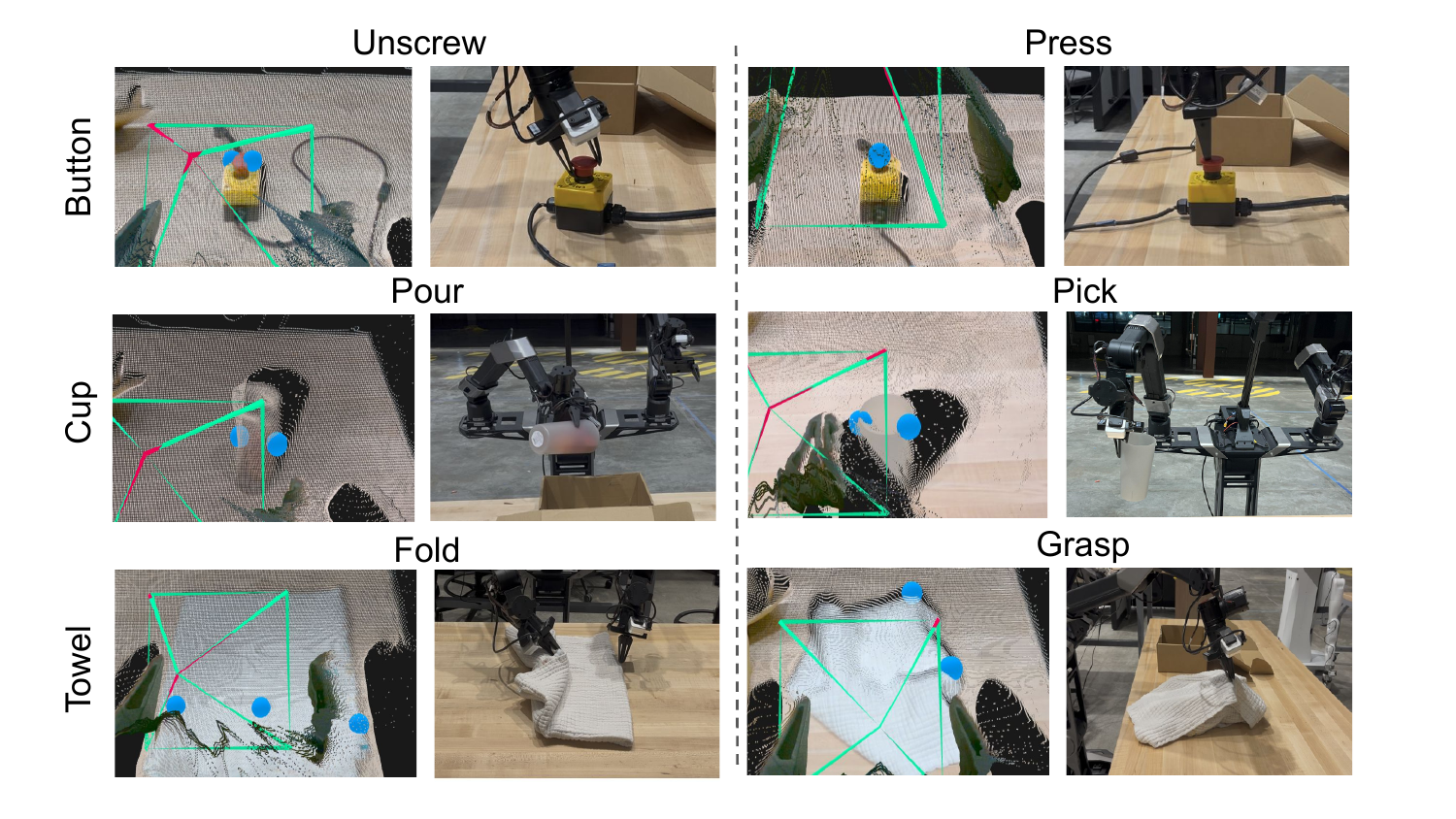}
    \caption{Skill-dependent geometric primitives for the same object. 
Left Column is the 3D GLB visualization from the 3D perception module; Right Column demonstrates the real skill execution performed by \SkillGraph. Even for same object, different skills require different semantic keypoints.}
    \label{fig:keypoints-variation}
    \vspace{-1em}
\end{figure}

\textbf{Effect of skill-specific grounding.}
As illustrated in \Cref{fig:keypoints-variation}, different skills applied to the same object may require different semantic keypoints.
Geometric contracts explicitly specify these object-centric requirements, guiding the promptable 3D perception module to identify interaction locations that satisfy both skill-level and task-level needs.
This explains why \method outperforms A1 (no contract) on three of the four tasks. MOKA underperforms on unscrewing and pouring because it cannot reliably generate the skill-specific rotational trajectories required by these tasks.
All methods achieve similar success rate on \emph{Press Button}, likely because the button center is relatively straightforward to localize and pressing requires only a short execution sequence, leaving limited room for skill-specific grounding to improve performance.

\begin{table}[t]
\centering
\small
\caption{Functional manipulation success rates (\%).}
\label{tab:functional}
\setlength{\tabcolsep}{6pt}
\begin{tabular}{@{}p{0.38\columnwidth}cccc@{}}
\toprule
Task & Ours & MOKA & A1 & A2 \\
\midrule
Unscrew button & \textbf{76.47} & 23.53 & 58.82 & 52.94 \\
Pour & \textbf{82.35} & 47.06 & 70.59 & 58.82 \\
Fold cloth edge to center & \textbf{64.71} & 58.82 & 35.29 & 52.94 \\
Press button & \underline{\textbf{76.47}} & \underline{\textbf{76.47}} & 70.59 & \underline{\textbf{76.47}} \\
\bottomrule
\end{tabular}
\vspace{-2em}
\end{table}

\begin{figure}[htbp]
    \centering
    \includegraphics[width=0.8\linewidth]{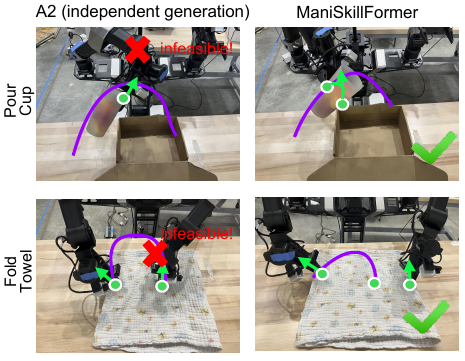}
    \caption{Comparison of A2 pipeline and \method. 
    }
    \label{fig:comparison-A2}
    \vspace{-2em}
\end{figure}

\textbf{Effect of joint generation.}
Compared with A2 (independent generation), \method improves by 23.53\% on \texttt{unscrew} and \texttt{pour}, and by 11.77\% on \texttt{fold}. As illustrated in \Cref{fig:comparison-A2}, geometric choices for one constituent skill must also support the requirements of subsequent skills. Jointly generating skills and contracts helps coordinate these choices within a meta skill, reducing mismatches between independently specified interaction geometry and execution requirements. The gains over A2 suggest that specifying a contract for each atomic skill is not always sufficient; compatibility between constituent skills also matters.  


\subsubsection{Q3 (Long-horizon performance)}

Except from composing simple atomic skills into complex skills, we can also compose skills (either atomic or meta) into long-horizon behabiviors. Table~\ref{tab:planning} evaluates long-horizon task performances, where errors may accumulate due to dependencies between successive skills. 
The tasks include repeated pick-and-place across six different objects, cone stacking requiring precise geometric alignment, and a multi-stage manipulation sequence with strong dependencies between successive skills.
The results demonstrate \method's ability to compose skills into long-horizon behaviors.
For simplicity, each step corresponds to an atomic skill, except for the meta skills \texttt{Pick} and \texttt{Place}, which are each counted as a single step.

\begin{table}[htbp]
\vspace{-0.8em}
\centering
\small
\caption{Long-horizon task success counts. Success requires completing the entire sequence without a failure.}
\label{tab:planning}
\setlength{\tabcolsep}{6pt}
\begin{tabular}{@{}p{0.45\columnwidth}cccc@{}}
\toprule
Task & Ours & MOKA & A1 & A2 \\
\midrule
Pick and Place 6 objects & \textbf{6/10} & 2/10 & 4/10 & \textbf{6/10} \\
Stack cones & \textbf{8/10} & 3/10 & 5/10 & \textbf{8/10} \\
Task C (11 steps) & \textbf{5/10} & 1/10 & 2/10 & 1/10 \\
\bottomrule
\end{tabular}
\vspace{-1em}
\end{table}

\textbf{Overall sequence completion.}
\method achieves the highest or joint-highest success count on all three tasks. A2 matches \method on the first task and cone stacking tasks because for each step (\texttt{Pick} or \texttt{Place}), they do not require much geometry dependencies. And \method outperforms A1 (no contract) ablation on cone stacking task because this task requires precise geometry estimation, which can be improved by the explicit contract. In general, these results suggest that explicit contracts support execution across the evaluated long-horizon sequences.

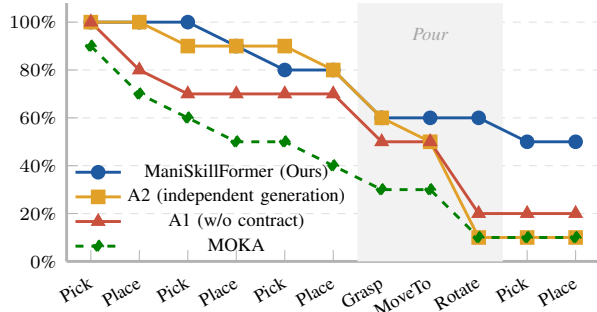
\begin{figure}[t]
\centering
\begin{tikzpicture}
\begin{axis}[
   width=1.0\linewidth,
    height=5.0cm,
    ymin=0, ymax=108,
    xmin=0.5, xmax=11.5,
    xtick={1,...,11},
    ytick={0,20,40,60,80,100},
    yticklabel={\pgfmathprintnumber{\tick}\%},
    xticklabels={Pick, Place, Pick, Place, Pick, Place,
                 Grasp, MoveTo, Rotate, Pick, Place},
    x tick label style={font=\scriptsize, rotate=30, anchor=north east},
    y tick label style={font=\scriptsize},
    axis line style={gray!60},
    tick style={gray!60},
    axis x line*=bottom,
    axis y line*=left,
    ymajorgrids=true,
    grid style={gray!25, dashed, line width=0.3pt},
    legend style={
    at={(0.55,-0.02)}, anchor=south east,
    draw=none, fill=none,
    font=\scriptsize, row sep=-1pt
    },
    line width=1.1pt,
    mark size=2.2pt,
    clip=false
]
\addplot[draw=none, fill=gray!10, forget plot]
    coordinates {(6.5,0) (9.5,0) (9.5,108) (6.5,108)} \closedcycle;
\node[font=\scriptsize\itshape, gray!70, anchor=south]
    at (axis cs:8,88) {Pour};
    
\addplot[color=oursblue, mark=*, mark options={fill=oursblue, draw=oursblue}]
    coordinates {(1,100) (2,100) (3,100) (4,90) (5,80) (6,80)
                 (7,60) (8,60) (9,60) (10,50) (11,50)};
\addlegendentry{\method (Ours)}

\addplot[color=a2gold, mark=square*, mark options={fill=a2gold, draw=a2gold}]
    coordinates {(1,100) (2,100) (3,90) (4,90) (5,90) (6,80)
                 (7,60) (8,50) (9,10) (10,10) (11,10)};
\addlegendentry{A2 (independent generation)}

\addplot[color=a1red, mark=triangle*, mark options={fill=a1red, draw=a1red}]
    coordinates {(1,100) (2,80) (3,70) (4,70) (5,70) (6,70)
                 (7,50) (8,50) (9,20) (10,20) (11,20)};
\addlegendentry{A1 (w/o contract)}

\addplot[
    color=green!50!black,
    dashed,
    mark=diamond*,
    mark options={fill=green!50!black, draw=green!50!black}
]
coordinates {
    (1,90) (2,70) (3,60) (4,50) (5,50) (6,40)
    (7,30) (8,30) (9,10) (10,10) (11,10)
};
\addlegendentry{MOKA}

\end{axis}
\end{tikzpicture}
\vspace{-1em}
\caption{Average survival rate over a 11-step long-horizon task C (``Pick yellow lego and blue lego into cup,  pick red lego into red bowl, pour cup into the box, pick red bowl and place into box"). Each step corresponds to a skill. \method with geometric contracts maintains higher survival rates and slows the accumulation of execution errors.}
\label{fig:survival_curve}
\vspace{-2em}
\end{figure}

\textbf{Reliability under dependency.}
Figure~\ref{fig:survival_curve} plots survival rate after each step in Task C (11 steps).
All methods start with comparably high success rates, but the MOKA baseline and the ablated pipelines declines sharply as errors accumulate, while \method maintains higher survival rates throughout, indicating that contract-based grounding produces more reliable intermediate states rather than merely more reliable individual steps.

The difference becomes evident in steps 7-9, corresponding to the meta skill \emph{Pour}, which decomposes into the atomic skills \emph{Grasp}, \emph{MoveTo}, and \emph{Rotate}. 
Successful ``pouring" requires ``grasping" the cup along two opposite sides of the cup body to provide stable support during ``rotation". 
If the grasp is placed on edge points with a vertical direction instead, the subsequent \emph{Rotate} skill cannot generate the correct pouring motion. 
Without geometric contracts, the ablated pipeline frequently selects such future-unaware grasp locations, which leads to a sharp drop in success rate during the \emph{Rotate} stage. 

\subsection{LLM Token Usage}
Offline, the contract and skill agents consume approximately
1,608 and 2,525 tokens per skill--object request pair,
respectively; their outputs are reused across executions.
Online, 2D keypoint prediction
uses approximately 1,518 tokens per request with a contract
and 1,429 without.
MOKA uses approximately 1,255 tokens per task.

\section{CONCLUSION AND FUTURE WORK}
\label{sec:conclusion}

We presented \method, motivated by the idea that reusable manipulation skills can be specified semantically before their geometry is grounded in a particular scene. Starting from human-defined skill schemas, LLM agents adapt motion templates to skill--object pairs and their task context, while geometric contracts make explicit what perception must provide for execution. These contracts then guide promptable 3D perception to recover the interaction geometry needed to turn the templates into robot motions. Experiments on the Galaxea R1-Lite show how this connection supports generalization across object instances, functional manipulation, and long-horizon composition.

For limitations, the current framework is limited to a predefined skill vocabulary and does not explicitly address environmental collision avoidance, dexterous manipulation, or compliance control.
Future work will integrate collision-aware motion planning and richer control strategies, and investigate skill abstractions that at the same time incorporate human knowledge while adapting to broader objects and tasks and also do automatic skill discovery.
The modular interface also provides a path to incorporating advances in multimodal perception (like GPT-6) while retaining reusable skill definitions.

\section{ACKNOWLEDGEMENT OF AI TOOL USAGE}
GPT-5.6 and Claude assisted with text polishing of the manuscript. All AI revisions were reviewed and verified by the authors, who take full responsibility for the final content.

{\small
\bibliographystyle{IEEEtran}
\bibliography{references}
}

\end{document}